\documentclass[a4paper]{spie}  %>>> use for US letter paper
\usepackage{amsmath,amsfonts,amssymb}
\usepackage{graphicx}
\usepackage[colorlinks=true, allcolors=blue]{hyperref}

\title{Machine learning-based colon deformation estimation method for colonoscope tracking}

\author[a]{Masahiro Oda}
\author[b]{Takayuki Kitasaka}
\author[c]{Kazuhiro Furukawa}
\author[d]{Ryoji Miyahara}
\author[c]{Yoshiki Hirooka}
\author[d]{Hidemi Goto}
\author[e]{Nassir Navab}
\author[a]{Kensaku Mori}
\affil[a]{Graduate School of Informatics, Nagoya University, Furo-cho, Chikusa-ku, Nagoya, Aichi, Japan}
\affil[b]{School of Information Science, Aichi Institute of Technology, 1247 Yachigusa, Yagusa-cho, Toyota, Aichi, Japan}
\affil[c]{Department of Endoscopy, Nagoya University Hospital, 65 Tsurumai-cho, Syouwa-ku, Nagoya, Aichi, Japan}
\affil[d]{Dept. of Gastro. and Hepato., Nagoya University Graduate School of Medicine, 1247 Yachigusa, Yagusa-cho, Toyota, Aichi, Japan}
\affil[e]{Technical University of Munich, Boltzmannstr. 3, 85748 Garching bei M\"{u}nchen, Germany}

\authorinfo{Further author information: 
(Send correspondence to M. Oda)
\\M. Oda: E-mail: moda@mori.m.is.nagoya-u.ac.jp, Telephone: +81 (0)52 789 5688
\\  T. Kitasaka: E-mail: kitasaka@aitech.ac.jp, Telephone: +81 (0)565 48 8121\\ K.Mori: E-mail: kensaku@is.nagoya-u.ac.jp, Telephone: +81 (0)52 789 5689}

\begin{document} 
\maketitle

\begin{abstract}
This paper presents a colon deformation estimation method, which can be used to estimate colon deformations during colonoscope insertions.
%This model contributes accuracy improvement of colonoscope tracking methods.
Colonoscope tracking or navigation system that navigates a physician to polyp positions during a colonoscope insertion is required to reduce complications such as colon perforation.
A previous colonoscope tracking method obtains a colonoscope position in the colon by registering a colonoscope shape and a colon shape.
The colonoscope shape is obtained using an electromagnetic sensor, and the colon shape is obtained from a CT volume.
However, large tracking errors were observed due to colon deformations occurred during colonoscope insertions.
Such deformations make the registration difficult.
Because the colon deformation is caused by a colonoscope, there is a strong relationship between the colon deformation and the colonoscope shape.
An estimation method of colon deformations occur during colonoscope insertions is necessary to reduce tracking errors.
We propose a colon deformation estimation method.
This method is used to estimate a deformed colon shape from a colonoscope shape.
%These two shapes are represented as sets of points.
We use the regression forests algorithm to estimate a deformed colon shape.
The regression forests algorithm is trained using pairs of colon and colonoscope shapes, which contains deformations occur during colonoscope insertions.
As a preliminary study, we utilized the method to estimate deformations of a colon phantom.
In our experiments, the proposed method correctly estimated deformed colon phantom shapes.
%The model can be applied to estimate real colon deformations if we have training data.
\end{abstract}

% Include a list of keywords after the abstract 
\keywords{Colon, colonoscope tracking, deformation estimation}

\section{INTRODUCTION}
\label{sec:intro}

%Purpose of this paper is to present a colon deformation estimation method during colonoscope insertions.
In the colon diagnosis and treatment, finding colonic polyps or early-stage cancers is important.
Most of colonic polyps are benign, but they may change to cancer.
Therefore, physicians try to find colonic polyps in colon diagnosis.
CT colonography is becoming a popular procedure as a new colon diagnostic method.
Physicians observe the inside of the colon on a CT volume of a patient in CT colonography.
If colonic polyps or early-stage cancers are found in diagnosis, a colonoscopic examination or polypectomy is performed.
A colonoscopic polypectomy is the removal of polyps.
A physician inserts a colonoscope to explore and find polyps in the colon.
Currently, colonoscope insertion is performed based on information obtained from colonoscopic images.
However, the field of view of a colonoscope is limited due to occlusions caused by many haustral folds and colonic wall.
A physician has to estimate the colonoscope position inside the colon based on her/his experience.
However, colon shape changes significantly during a colonoscope insertion.
Inexperienced physicians may cause complications such as colon perforation.
A colonoscope navigation system that navigates a physician to polyp positions during a colonoscope insertion is necessary to reduce such problems.

A colonoscope tracking method is one of fundamental function in the navigation systems.
It estimates the colonoscope position in the colon.
As related work of colonoscope tracking, many bronchoscope tracking methods have been proposed.
These methods include image-based and sensor-based ones\cite{Peters08,Deligianni05,Rai08,Deguchi09,Gildea06,Schwarz06}.
Previous colonoscope tracking methods also include image-based and sensor-based methods.
Although image-based colonoscope tracking method is proposed\cite{Liu13}, it is difficult to keep tracking from unclear colonoscopic views.
Electromagnetic (EM) sensors can be used to obtain positions in the colon\cite{Ching10,Fukuzawa15}.
A colonoscope tracking method that uses a CT volume and an electromagnetic (EM) sensor was reported\cite{Oda17}.
This method obtains two curved lines representing colon and colonoscope shapes to estimate a colonoscope position on a CT volume coordinate system.
They can track the colonoscope position regardless of the colonoscopic image quality.
However, colon deformations occurred during the colonoscope insertions are not considered.
The line registration does not work well when the colon deforms.
It resulted in large errors of the colonoscope tracking.

To improve the tracking accuracy, we propose an estimation method of the colon deformation from colonoscope shape.
This is because there is a strong relationship between the colon deformation and the colonoscope shape.
We can obtain the colonoscope shape from the EM sensors.
The colonoscope shape is an important information to estimate colon deformations.
In this paper, we develop a colon deformation estimation method, which can be used to estimate a colon shape or deformation from a colonoscope shape.
This method establishes correspondences between colonoscope shapes and colon shapes, that are captured during colonoscope insertions.
By using this method, we can obtain a deformed colon shape from a colonoscope shape measured by EM sensors.
The deformed colon shape can be used to estimate a colonoscope position in the colon during a colonoscope tracking.

We present the colon deformation estimation method in Section \ref{sec:estimation}.
As an application example of the method, we tried estimation of colon phantom deformations in Section \ref{sec:experiments}.

\section{COLON DEFORMATION ESTIMATION BY REGRESSION FORESTS}
\label{sec:estimation}

\subsection{Overview}

We estimate colon shape deformation from a colonoscope shape using the regression forests method.
To perform this estimation, we need pairs of a colonoscope shape and a colon shape captured during a colonoscope insertion.
The regression forests are trained using the shape pairs.
After training, the regressors become possible to estimate a colon shape from a colonoscope shape.

\subsection{Shape representations}

There are many methods in 3D shape representations, such as surface mesh, volumetric, centerline, and point set representations.
We employ a point set representation to describe the colonoscope and colon shape.
The colonoscope and colon shapes are represented as sets of points aligned along their centerlines.
Because both of colonoscope and colon shapes are no-branching strap-shaped, they are suitable for the point set representation.

%Both of these shapes are represented as sets of points.
The colonoscope shape is a set of points ${\bf X}=({\bf x}_{1}, \ldots, {\bf x}_{N})$ aligned along a colonoscope.
$N$ is the total number of points contained in the colonoscope shape.
The colon shape is a set of points ${\bf Y}=({\bf y}_{1}, \ldots, {\bf y}_{m}, \ldots, {\bf y}_{M})$ aligned along a colon centerline.
$M$ and $m$ are the total number of points contained in the colon shape and the index of the points, respectively.

\subsection{Estimation}

\begin{figure}[tb]
\begin{center}
\includegraphics[width=0.5\textwidth]{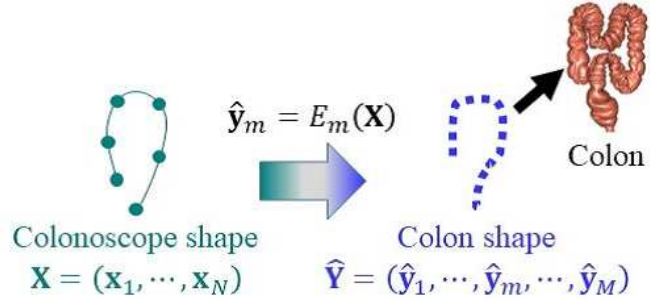}
\caption{Estimation of colon shape from colonoscope shape.}
\label{fig:estimation}
\end{center}
\end{figure}

The regression forests method is used to obtain an estimated point of ${\bf y}_{m}$, which is denoted as $\hat{{\bf y}}_{m}$, from a colonoscope shape ${\bf X}$.
This estimation is illustrated in Fig. \ref{fig:estimation}.
We describe a regressor for ${\bf y}_{m}$ as 
\begin{equation}
\hat{{\bf y}}_{m} = E_{m}({\bf X}),
\end{equation}
where $E_{m}$ is a trained regression forests regressor.
We train $M$ regressors using pairs of ${\bf X}$ and ${\bf Y}$ in training data.
By training the regressors using training data containing deformed colon shapes, the regressors can estimate deformed colon shapes.
In each regressors, the number of trees in the forest is set as 100.

In our colon deformation estimation, we input a colonoscope shape ${\bf X}$ to the regressors.
The regressors  calculates an estimated colon shape $\hat{{\bf Y}}=(\hat{{\bf y}}_{1}, \ldots, \hat{{\bf y}}_{M})$.

%The colon deformation model defines relationships between colonoscope and colon shapes.
%This model is used to obtain an estimated colon shape $\hat{{\bf Y}}$ from a colonoscope shape ${\bf X}$, which is measured during a colonoscope insertion.
%This model $E$ is represented as
%\begin{equation}
%\hat{{\bf Y}} = E({\bf X}).
%\end{equation}
%This is regression from an explanatory variable $X$ to a target variable $Y$.
%We use the regression forests method as the colon deformation model.
%Because the regression forests method is a machine learning-based regression approach, we measure training data of the colonoscope and colon shapes for training of it.
%The regression forests are trained using training data.
%After a training process, the colon deformation model is established.

\section{EXPERIMENTS}
\label{sec:experiments}

\subsection{Materials}

To evaluate our colon shape deformation estimation method, we have conducted phantom-based experiments.
We tested the proposed method in a colon phantom study.
%We tried to estimate colon deformations using a colon phantom.
%As a preliminary study, we build a colon deformation model of a colon phantom, then we try to estimate deformations using the model.
%We used a colon phantom because measurement of colon shapes during colonoscope insertions is easy.
This experiment measures colon shapes of the phantom by a distance sensor and colonoscope shapes by an EM sensor.
%Relationships between these shapes are established using the colon deformation model.
%Colon deformation estimation was performed and visually evaluated.
%Of course, this model can be applied for estimations of real colon deformations.
%To estimate real colon deformations, we will measure the shapes of real colons by using X-ray imaging techniques while colonoscope insertions.
We used a colon phantom (Colonoscopy training model type I-B, Koken, Tokyo, Japan), a CT volume of the phantom, a colonoscope (CF-Q260AI, Olympus, Tokyo, Japan), an EM sensor (Aurora Shape Tool Type 1, NDI, Ontario, Canada), and a distance image sensor (Kinect v2, Microsoft, WA, USA).
We measured colonoscope and colon shapes during seven times of colonoscope insertions as explained in Sections \ref{ssec:colonoscopemeasure} and \ref{ssec:colonmeasure}.
The colonoscope was moved from the cecum to the anus.
An engineering researcher operated the colonoscope.
%We measured 134 frames of colonoscope and colon shapes pairs per one colonoscope insertion.
We train the regression forests using frames of six times colonoscope insertions, then, we perform a colon deformation estimation using colonoscope shapes contained in frames of the remaining one colonoscope insertion.
Details of the training and estimation processes are explained in Sections \ref{ssec:training} and \ref{ssec:estimation}.

\subsection{Colonoscope shape measurement} \label{ssec:colonoscopemeasure}

We used the colonoscope to measure its shape.
We inserted an EM sensor, which has six EM sensors, to the working channel of the colonoscope.
The measured data is a set of six three-dimensional points ${\bf X}=({\bf x}_{1}, \ldots, {\bf x}_{6})$ aligned along the colonoscope and is considered as a colonoscope shape.
%This is a colonoscope shape.

\subsection{Colon shape measurement} \label{ssec:colonmeasure}

%We used a colon phantom (Colonoscopy training model type I-B, Koken, Tokyo, Japan).
The distance image sensor is mounted to measure the surface shape of the colon phantom.
We attached 12 position markers on the surface of the colon phantom.
Both of a distance and color images are obtained from the distance image sensor.
We applied an automated marker position extraction process to these images to obtain 12 three-dimensional points of the markers.
The measured marker points were aligned along the colon centerline and numbered.
The numbered markers are described as ${\bf Y}=({\bf y}_{1}, \ldots, {\bf y}_{m}, \ldots, {\bf y}_{12})$.
$m$ is the index of the marker.
${\bf y}_{1}$ and ${\bf y}_{12}$ correspond to a marker near the cecum and a marker near the anus, respectively.
${\bf Y}$ is a colon shape of the phantom.

\subsection{Regression forests training} \label{ssec:training}

\begin{figure}[tb]
\begin{center}
\includegraphics[width=0.9\textwidth]{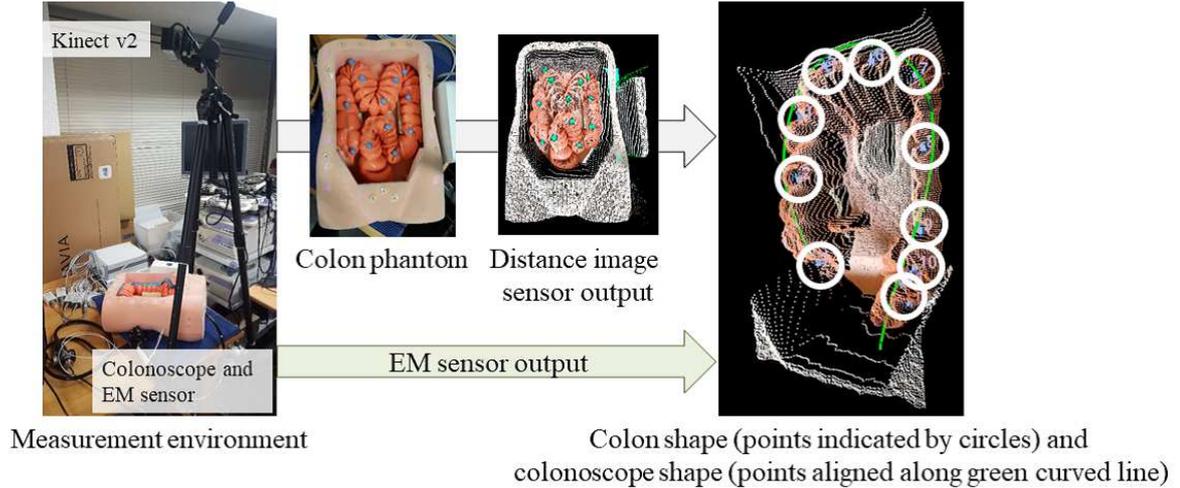}
\caption{Colonoscope and colon shapes measurement. Sensors are mounted as shown in left figure. Sensor outputs are registered as shown in right figure.}
\label{fig:measurement}
\end{center}
\end{figure}

During colonoscope insertions to the colon phantom, we measured both of ${\bf X}^{t}$ and ${\bf Y}^{t} \ (t=1,\ldots,T)$ simultaneously.
The measurement environment is shown in Fig. \ref{fig:measurement}.
The measurement was performed six times per second.
$t$ and $T$ mean the index of measurement time and the total data number.
Wrong measurement results caused by mis-detection of markers are manually corrected.
The ${\bf X}^{t}$ and ${\bf Y}^{t}$ contain points in the EM and distance sensors coordinate systems, respectively.
We registered them into the CT volume coordinate system using the iterative closest point (ICP) algorithm and a manual registration.
We trained regression forests of each marker using ${\bf X}^{t}$ and ${\bf y}^{t}_{m} \  (t=1,\ldots,T)$.
Totally, 12 regressors were built.

\subsection{Colon deformation estimation} \label{ssec:estimation}

We estimated colon shape or deformation using the regressors.
We measured a current colonoscope shape ${\bf X}$ while it was inserted to the colon.
${\bf X}$ was transformed to the CT coordinate system using the ICP algorithm.
We obtained an estimated colon shape $\hat{{\bf Y}}$ using the 12 regressors.

The estimated $\hat{{\bf Y}}$ represents marker positions on the colon surface.
We apply smoothing along time-series for each point in $\hat{{\bf Y}}$ to make smooth deformation.
The smoothing result is used to estimate a colon shape or displacements of parts of the colon during a colonoscope insertion.

\section{RESULTS}
\label{sec:results}

\begin{figure}[tb]
\begin{center}
\begin{tabular}{ccc}
\includegraphics[width=0.3\textwidth]{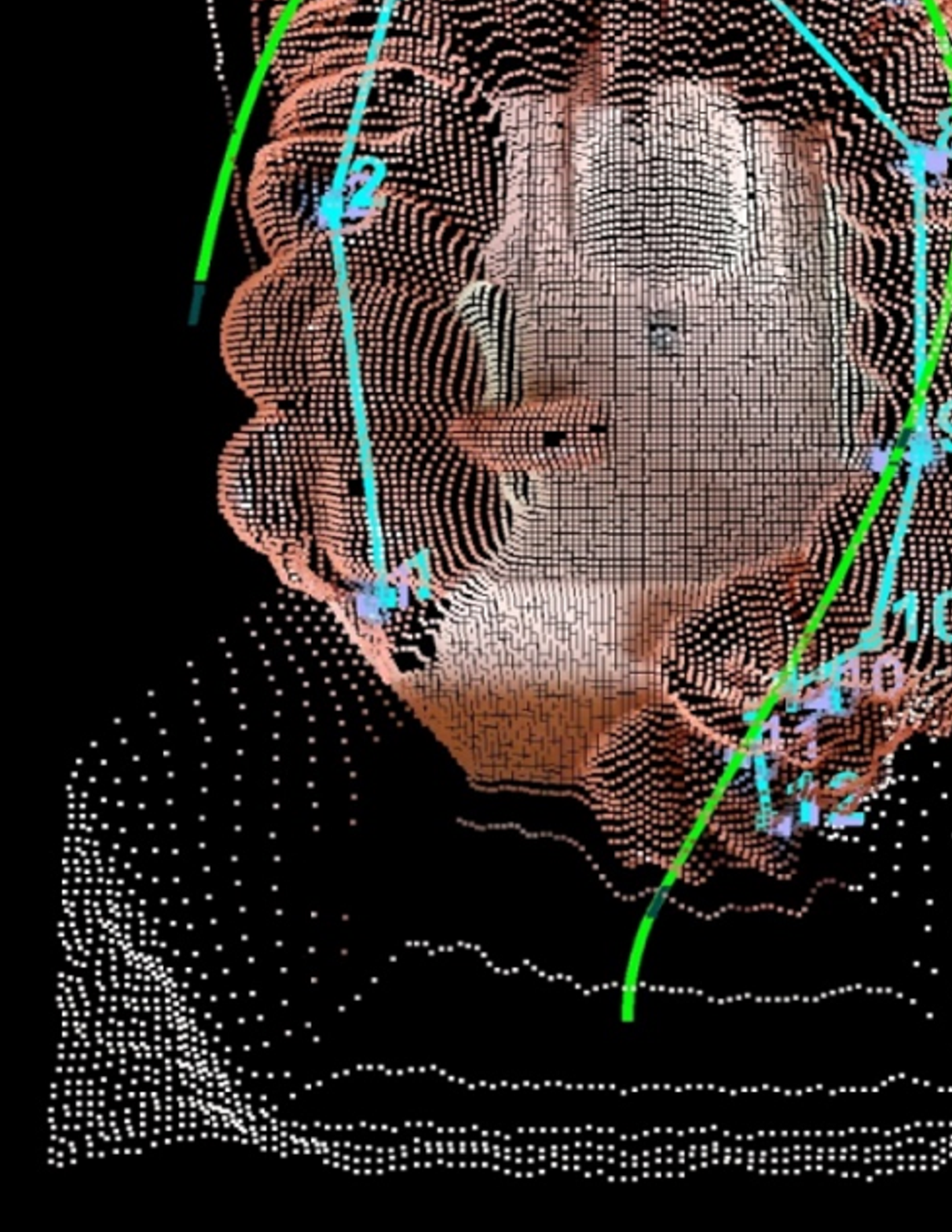} & 
\includegraphics[width=0.295\textwidth]{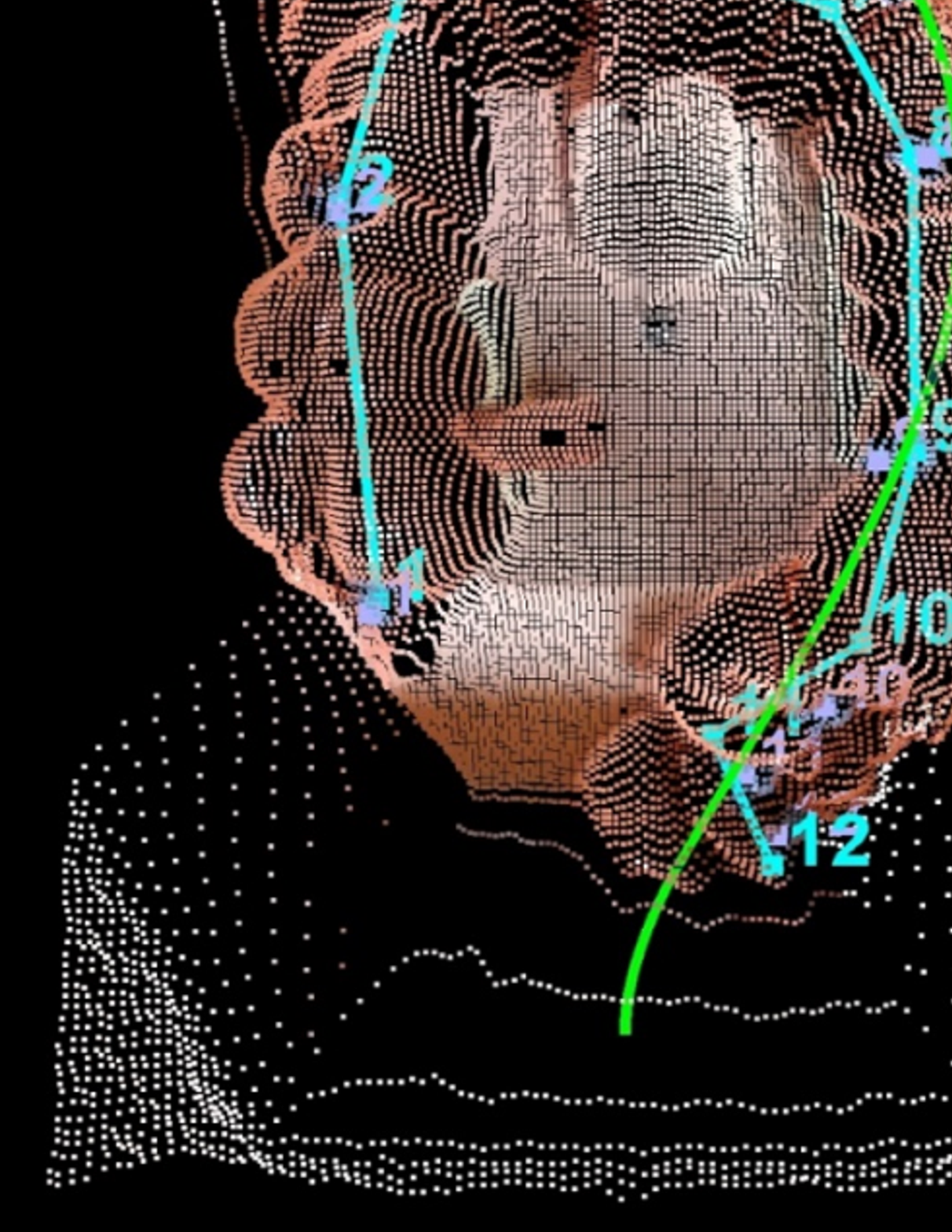} & 
\includegraphics[width=0.268\textwidth]{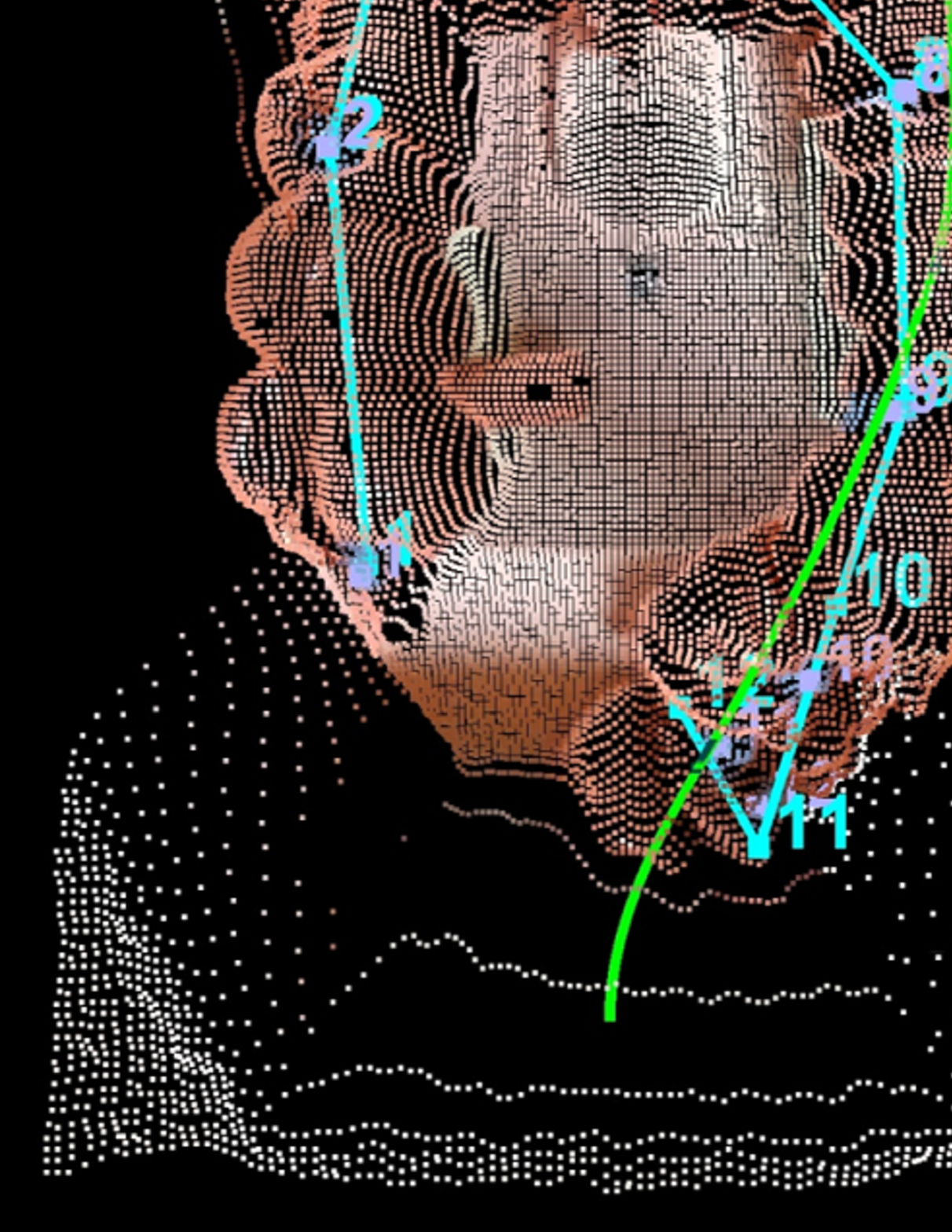}\\
(a) & (b) & (c)\\
\includegraphics[width=0.3\textwidth]{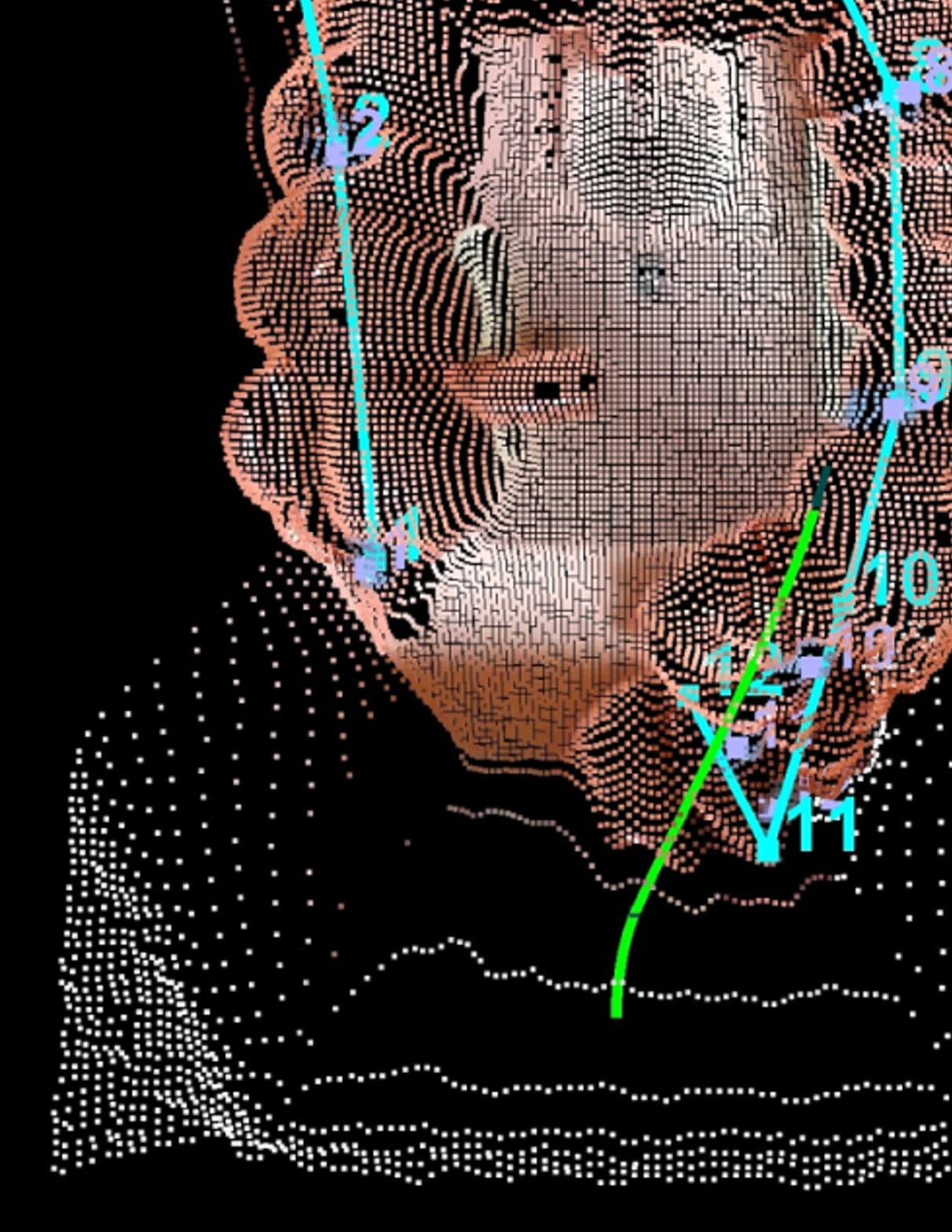} & 
\includegraphics[width=0.295\textwidth]{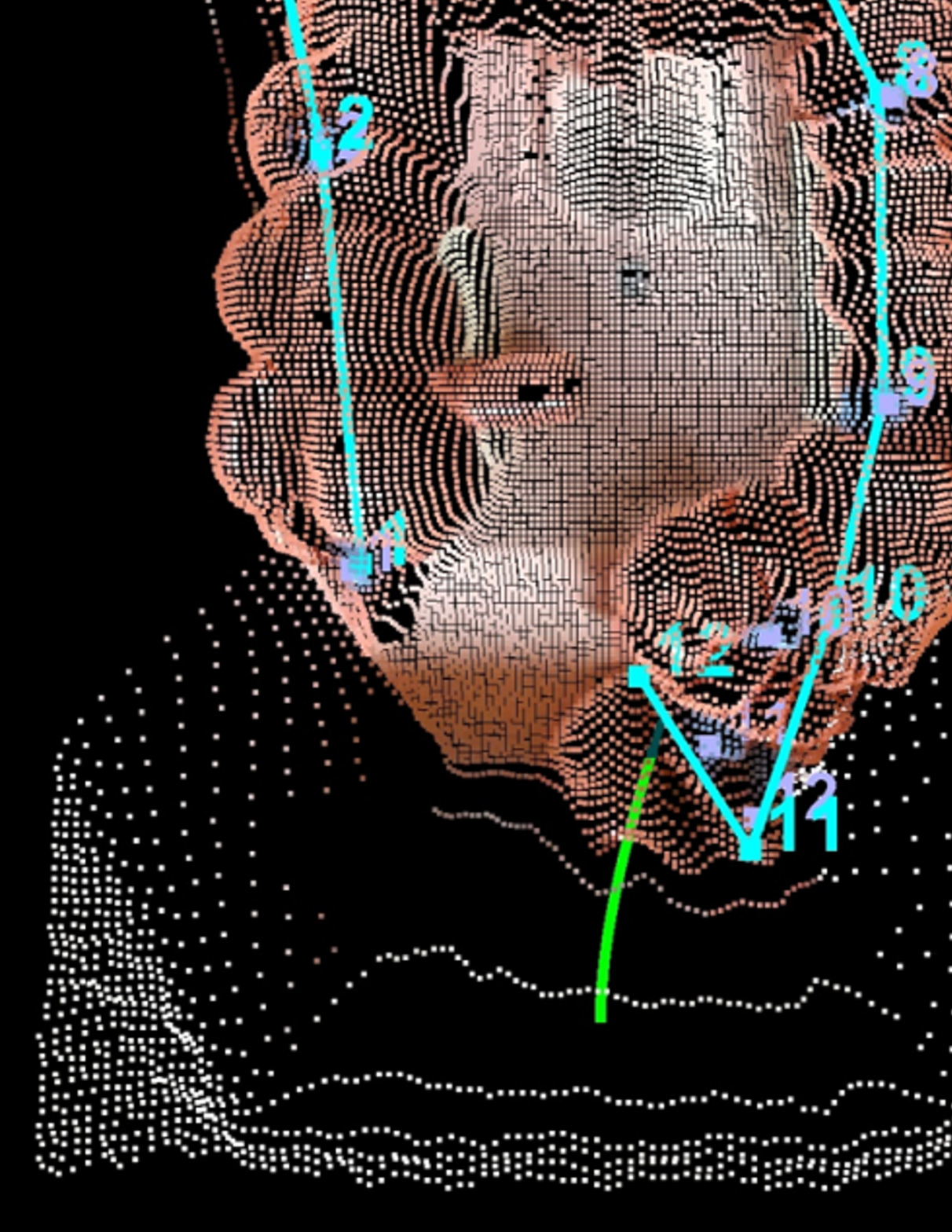}\\
(d) & (e)
\end{tabular}
\caption{Colonoscope shapes (points aligned along green curved lines), surface shapes of colon phantom (dots), and estimated colon shapes (points connected by sky blue polygonal lines). Colonoscope was inserted to near cecum (a) to anus (e),}
\label{fig:result}
\end{center}
\end{figure}

Estimated colon shape (marker positions) is shown in Fig. \ref{fig:result}.
This figure shows colonoscope shapes (points aligned along green curved lines), surface shapes of the colon phantom (dots), and estimated colon shapes (points connected by sky blue polygonal lines) at five colonoscope insertion states.
The estimated colon shapes were located on the colon phantom.
It indicates the proposed colon deformation estimation method can estimate colon shapes at various colonoscope insertion states.

\section{DISCUSSION}
\label{sec:discussion}

The method performs estimation during colonoscope movements from the cecum to the anus.
By using this estimation results, deformed colon shapes are easily obtained.
The deformed colon shapes can be used to show colon shapes in colon navigation systems.
Also, the deformed colon shapes are utilized to track colonoscope position using a tracking method\cite{Oda17}.
We believe that the proposed method contributes accuracy improvements of colonoscope tracking.

Some points on the estimated colon shape were located outside of the colon surface.
We performed estimation of point independently.
Points on the colon have positional relationships.
They change positions under physical constraints related to the colon phantom.
We need to take the physical constraints into account in the colon deformation estimation process.
Our estimation result will be improved by considering physical properties.
%Wrong estimation results were observed sometimes.
%We found the automated marker position extraction process in the colon shape measurement caused wrong detection or overlooking sometimes.
%The measured colon shapes are used in training of the method.
%Wrong detection or overlooking of markers reduces accuracy of the estimation.
%The automated marker position extraction process needs to be improved to obtain more accurate estimation results from the method.

\section{CONCLUSIONS}
\label{sec:conclusions}

%We presented the colon deformation model, which contributes improvements of colonoscope tracking accuracies.
We presented a colon deformation estimation method and its application to a colon phantom.
Pairs of colonoscope and colon shapes during colonoscope insertions are measured by an EM sensor and a distance image sensor.
The estimation method, which consists of regression forests, are trained to estimate a colon shape from a colonoscope shape.
In our experiments using a colon phantom, the method estimated colon shapes during colonoscope insertions.
Future work includes improvement of the automated marker position extraction process, measurements of colonoscope and colon shapes under physician's operations, and development of a colonoscope navigation system.

\acknowledgments % equivalent to \section*{ACKNOWLEDGMENTS}       
 
Parts of this research were supported by the MEXT/JSPS KAKENHI Grant Numbers 25242047, 26108006, 17H00867, the JSPS Bilateral International Collaboration Grants, and the JST ACT-I (JPMJPR16U9).

% References
\bibliography{18spie_paper_cite} % bibliography data in report.bib
\bibliographystyle{spiebib} % makes bibtex use spiebib.bst

\end{document}